\documentclass{article}
\usepackage{spconf,amsmath, amssymb, graphicx}
\usepackage{hyperref} 
\usepackage{graphicx}
\usepackage{xcolor}
\usepackage{algorithm2e}
\usepackage{algorithmicx}
\usepackage{setspace}
\usepackage[export]{adjustbox}
\usepackage{multirow}
\usepackage[noend]{algpseudocode}

\DeclareMathOperator*{\argmin}{arg\,min}

\title{Subset Sampling For Progressive Neural Network Learning}
\name{Dat Thanh Tran$^{1}$, Moncef Gabbouj$^{1}$, Alexandros Iosifidis$^{2}$}
\address{$^{1}$Department of Computing Sciences, Tampere University, Finland\\
	$^{2}$Department of Engineering, Aarhus University, Denmark\\
	Emails: \{thanh.tran, moncef.gabbouj\}@tuni.fi, alexandros.iosifidis@eng.au.dk}

\begin{document}
%
\maketitle
\begin{abstract}
Progressive Neural Network Learning is a class of algorithms that incrementally construct the network's topology and optimize its parameters based on the training data. While this approach exempts the users from the manual task of designing and validating multiple network topologies, it often requires an enormous number of computations. In this paper, we propose to speed up this process by exploiting subsets of training data at each incremental training step. Three different sampling strategies for selecting the training samples according to different criteria are proposed and evaluated. We also propose to perform online hyperparameter selection during the network progression, which further reduces the overall training time. Experimental results in object, scene and face recognition problems demonstrate that the proposed approach speeds up the optimization procedure considerably while operating on par with the baseline approach exploiting the entire training set throughout the training process. 

\end{abstract}

\section{Introduction}
Progressive Neural Network Learning (PNNL) \cite{huang2007convex, zhou2014stacked, chatterjee2017progressive, chen2017broad, kiranyaz2017progressive, tran2018progressive, tran2019heterogeneous, tran2019learning} aims to build the network's topology incrementally depending on the training set given for the specific problem at hand. At each incremental training step, a PNNL algorithm adds a new set of neurons to the existing network topology and optimizes the new synaptic weights using the entire training set. Thus, throughout the network's topology progression, the number of times the PNNL algorithm iterates through the entire training set is very high. For large datasets, this approach leads to an enormous computational cost and long training process. In this paper, we propose to perform the optimization of each incremental training step using only a subset of the training data. Our motivation in doing so is two-fold. Firstly, optimizing with respect to a subset of the training data leads to lower overall computational cost; Secondly, the use of different subsets of data at each incremental step promotes specialization of different sets of neurons at capturing different patterns in the data.

The idea of subset sampling for training machine learning methods has been proposed in different contexts in literature. With the motivation of reducing the expense of labeling data for training, methods following the active learning paradigm \cite{settles2009active} seek to define a sampling strategy that selects a sample to be labeled among a large pool of unlabeled data for the next learning round. While the active learning paradigm considers the problem of data selection in an (initially) unsupervised setting, in the context of PNNL we take advantage of the available labeling information for subset selection. Directly related to our work are methods selecting a subset of data formed by the most representative samples \cite{sener2017active, birodkar2019semantic, shokri2019learning, banerjee2019deepsub}. These methods however only consider the data selection process once based on the input data representations and the available labels. The selected subset of data is then used to train a model with fixed capacity. Different from this line of works, we propose to perform subset sampling at every incremental step of the PNNL process with selection strategies that can also take into account the data representations learned by the current network's tolpology. 

When building a learning system, the development process often requires running multiple experiments to select the best values for the hyper-parameters associated with the learning model. For neural networks, such hyper-parameters correspond to the values used e.g. for the weight decay coefficient, or the dropout percentage. In existing PNNL algorithms, the value associated with each hyper-parameter is fixed throughout the entire training process, and the best combination of the hyper-parameter values is usually selected by following a grid search strategy training multiple models each corresponding to a different combination of hyper-parameter values. Different from that (traditional) approach, we propose to incorporate the hyper-parameter selection process into each incremental training step, enabling adaptive hyper-parameter assignment during the network's topology progression process. Coupled with the speed up gained from subset sampling, this further accelerates the overall training process and improves generalization performance as indicated by our experimental results. 
  
The remainder of the paper is organized as follows: Section 2 reviews Progressive Neural Network Learning and the subset sampling strategies in different learning contexts. Section 3 describes the proposed progressive network training method. In Section 4, we detail our experimental setup and present empirical results. Section 5 concludes our work.

\section{Related works}
\subsection{Progressive Neural Network Learning}
In Progressive Neural Network Learning (PNNL), an algorithm starts with an initial network topology and gradually increases the capacity of the model by adding and optimizing new blocks of neurons following an iterative optimization process \cite{huang2007convex, zhou2014stacked, chatterjee2017progressive, chen2017broad, kiranyaz2017progressive, tran2018progressive, tran2019heterogeneous, tran2019learning, tran2019data, tran2019knowledge}. When a new set of neurons is added to the current network topology, different PNNL algorithms determine different rules to form new synaptic connections from the new neurons to the existing ones. For example, in I-ELM \cite{huang2007convex} and BLS \cite{chen2017broad}, the progression strategies only allow the algorithms to learn networks with one and two hidden layers, respectively, while other PNNL algorithms such as PLN \cite{chatterjee2017progressive}, StackedELM \cite{zhou2014stacked} or HeMLGOP \cite{tran2019heterogeneous} can generate multilayer networks. 

Regarding the adopted optimization strategies, many algorithms employ random hidden neurons to relax the optimization objective to a convex form and use convex optimization techniques to achieve global solutions such as \cite{huang2007convex, chen2017broad, zhou2014stacked, chatterjee2017progressive}. While this approach is computationally efficient and often comes with certain theoretical guarantees, most algorithms are sensitive to hyper-parameter selection and require extensive evaluation of a large set of hyper-parameter values. Besides, these algorithms often construct very large network topologies to achieve good performance. Recently, the authors in \cite{tran2019heterogeneous} proposed HeMLGOP, a PNNL algorithm that combines both randomization process and stochastic optimization to progressively train networks of heterogeneous neurons. Since HeMLGOP not only optimizes the network's topology but also the functional form of each neuron, the resulting network are both compact and efficient. This, however, comes with a much higher training computational cost compared to those employing random neurons and convex optimization. 

As a variant of HeMLGOP algorithm which only optimizes the network's topology with the standard Perceptron, Progressive Multilayer Perceptron (PMLP) yields a good trade-off between optimization complexity, topology compactness and learning capability. Thus, in this paper we apply our proposal to speed-up and enhance PMLP. Although our investigation in this paper limits to only PMLP, the proposed method can be generalized to all PNNL algorithms as will be indicated in the next Section.

\subsection{Subset Sampling}
In Active Learning, query-acquiring or pool-based method refers to a class of algorithms that uses different sampling strategies to select the most informative samples from a pool of unlabeled data. The most representative examples in this category of methods include the information-theoretic method in \cite{mackay1992information}, the ensemble method in \cite{mccallumzy1998employing}, and the method based on uncertainty heuristics in \cite{tong2001support}. For a comprehensive review of active learning methods, we refer the reader to \cite{settles2009active}. 

Subset sampling methods have also been proposed in different contexts. For example, submodular function optimization for selecting a subset of samples was proposed in \cite{banerjee2019deepsub} to speed up neural network training. To study sample redundancy, \cite{birodkar2019semantic} performs clustering using representations generated by a pre-trained model, while \cite{vodrahalli2018all} measures the importance of a sample via the gradient information. In the context of dataset compression and distributed learning, \cite{shokri2019learning} optimizes sample selection and model's parameters iteratively based on convex optimization.  

\section{Proposed Method}
\subsection{Subset Sampling}
Let us denote by $\mathbf{T} = \{(\mathbf{x}_i, \mathbf{y}_i) | i=1, \dots, N\}$ the training set formed $N$ samples with $\mathbf{x}_i$ and $\mathbf{y}_i$ being the $i$-th sample and its label, respectively. Let us also denote by $f_k(\mathbf{x}, \Theta_k, \Lambda_k)$ the function induced by the neural network's topology at the progression step $k$, where $\Theta_k$ representing the set of parameters to optimize, and $\Lambda_k$ representing the set of hyper-parameters. 

At step $k$, instead of optimizing $f_k$ with respect to $\Theta_k$ on $\mathbf{T}$, we propose to solve the optimization problem on a subset $\mathbf{S}_k \subset \mathbf{T}$ having cardinality $M \ll N$, i.e.:

\begin{equation}\label{eq1}
\argmin_{\Theta_k} \sum_{(\mathbf{x}_j, \mathbf{y}_j) \in \mathbf{S}_k} \mathcal{L}\big(f_k(\mathbf{x}_j, \Theta_k, \Lambda_k), \mathbf{y}_j\big)
\end{equation}
where $\mathcal{L}$ denotes the loss function. To this end, we evaluate three different sample selection methods defined based on the following criteria:

\begin{itemize}
\item \textit{Random Sampling}: at each progression step $k$, we form $\mathbf{S}_k$ by uniformly selecting $M$ samples from $\mathbf{T}$. Although random sampling has been theoretically proven to be inferior to other sampling strategies in many learning contexts \cite{freund1997selective, gilad2006query, settles2009active, banerjee2019deepsub}, as it will be shown by our empirical study, this is not necessarily the case for PNNL. Throughout the architecture progression process, random sampling ensures diverse sets of samples being iteratively presented to the network, thus promoting diversity of the newly added neurons with respect to the existing ones. 

\item \textit{Top-M Sampling based on miss-classification}: at each progression step $k$, this method computes the loss induced by each sample in $\mathbf{T}$ using the network's topology learned at step $k-1$, i.e., $\mathcal{L}\big(f_{k-1}(\mathbf{x}_i), \mathbf{y}_i\big)$, and selects the top $M$ samples which induce the highest loss values. Since the loss values directly provide supervisory signal when updating the model's parameters, by conditioning on the current model's knowledge expressed via $f_{k-1}$, this strategy enforces a given algorithm to learn new blocks of neurons which can correctly classify the most difficult cases.

\item \textit{Top-M Sampling based on diverse miss-classification}: while the previous sampling method solely considers the most difficult to classify samples, this strategy also aims to promote diversity and reduce similarity among the selected samples. To do so, we perform K-Means clustering using $f_{k-1}(\mathbf{x}_i)$ as inputs. The number of clusters $C$, which is pre-defined, can be set using simple heuristics such as being equal to the number of classes in classification tasks. We also compute the loss value induced by each sample using $f_{k-1}$ and select the top $m$ samples that induce highest loss values for every cluster, with $m = \lfloor M/C \rfloor$. 
\end{itemize}

\subsection{Online Hyperparameter Selection}
\begin{table}[t!]
	\begin{center}
		\caption{Test accuracy (\%). bold-face results indicate the best performance in each column.}\label{t1}
		\resizebox{0.7\linewidth}{!}{
			\begin{tabular}{|c|c|c|c|}
				\multicolumn{4}{c}{} \\ \hline
				\textbf{Models}		& \textbf{Caltech256} 	& \textbf{MIT} & \textbf{CelebA} \\ \hline \hline
				\multicolumn{4}{|c|}{\textit{Subset Percentage 10\%}} \\ \hline
				PMLP-Random & $\mathbf{80.27}$ & $\mathbf{69.93}$ & $\mathbf{90.33}$ \\ \hline
				PMLP-Top-Loss & $73.08$ & $66.28$ & $82.68$ \\ \hline
				PMLP-C-Top-Loss & $77.92$ & $66.73$ & $84.67$ \\ \hline \hline
				\multicolumn{4}{|c|}{\textit{Subset Percentage 20\%}} \\ \hline
				PMLP-Random & $70.21$ & $61.87$ & $79.61$ \\ \hline
				PMLP-Top-Loss & $74.64$ & $66.13$ & $87.38$ \\ \hline
				PMLP-C-Top-loss & $72.16$ & $68.61$ & $81.66$ \\ \hline \hline
				\multicolumn{4}{|c|}{\textit{Subset Percentage 30\%}} \\ \hline
				PMLP-Random & $72.99$ & $63.66$ & $83.03$ \\ \hline
				PMLP-Top-Loss & $72.61$ & $65.64$ & $86.30$ \\ \hline
				PMLP-C-Top-loss & $72.63$ & $61.91$ & $84.98$ \\ \hline \hline 
				\multicolumn{4}{|c|}{\textit{Full Set}} \\ \hline
				PMLP & $79.48$ & $69.29$ & $87.99$ \\ \hline
				StackedELM \cite{zhou2014stacked} & $56.66$ & $61.69$ & $45.37$ \\ \hline
				PLN \cite{chatterjee2017progressive} & $78.29$ & $67.46$ & $87.82$ \\ \hline	
				
			\end{tabular}
		}
	\end{center}
\end{table}

In most existing PNNL algorithms, the value of each hyper-parameter is fixed throughout the network's topology progression. An algorithm is run for all combinations of hyper-parameter values defined a-priori, and the hyper-parameter values combination leading to the best performance on the validation set is selected for final model deployment. 

Since PNNL algorithms gradually increase the complexity of the neural network, it is intuitive that the model might require different degrees of regularization at different stages. Besides, with subset sampling incorporated, we train new blocks of neurons with different subsets of training samples at each step, which might require different hyperparameter configurations. Thus, instead of performing hyper-parameter selection in an \textit{offline} fashion, we propose to incorporate the hyper-parameter selection procedure into progressive learning at every incremental step. 

Particularly, let $\mathbf{H}$ be the set of all hyper-parameter values combinations, and $Q$ be the cardinality of $\mathbf{H}$. At each progression step $k$, after determining $\mathbf{S}_k$, we solve $Q$ optimization problems corresponding to $Q$ assignments of hyper-parameter values:

\begin{equation}\label{eq2}
\begin{split}
\Theta_k^h = & \argmin_{\Theta_k} \sum_{(\mathbf{x}_j, \mathbf{y}_j) \in \mathbf{S}_k} \mathcal{L}\big(f_k(\mathbf{x}_j, \Theta_k, \Lambda_k^h), \mathbf{y}_j\big) \\
& \forall \Lambda_k^h \in \mathbf{H}
\end{split}
\end{equation}

The algorithm then selects $\Theta_k^h$ that achieves the best performance on the validation set for the newly added block of neurons. Online selection not only ensures the best hyper-parameter values selection for each newly added block of neurons, but also reduces the computation overhead incurred when running $Q$ individual network progression steps. 

\section{Experiments}

\begin{table}[t!]
	\begin{center}
		\caption{Number of unique samples (in thousands) selected by each strategy during network topology construction}\label{t2}
		\resizebox{0.7\linewidth}{!}{
			\begin{tabular}{|c|c|c|c|}
				\multicolumn{4}{c}{} \\ \hline
				\textbf{Models}		& \textbf{Caltech256} 	& \textbf{MIT} & \textbf{CelebA} \\ \hline \hline
				\multicolumn{4}{|c|}{\textit{Subset Percentage 10\%}} \\ \hline
				PMLP-Random & $16.5$ & $7.8$ & $41.7$ \\ \hline 
				PMLP-Top-Loss & $8.3$ & $4.3$ & $16.1$ \\ \hline 
				PMLP-C-Top-Loss & $12.0$ & $4.0$ & $18.4$ \\ \hline \hline
				\multicolumn{4}{|c|}{\textit{Subset Percentage 20\%}} \\ \hline
				PMLP-Random & $22.4$ & $10.9$ & $46.6$ \\ \hline 
				PMLP-Top-Loss & $14.0$ & $6.8$ & $31.1$ \\ \hline 
				PMLP-C-Top-loss & $18.0$ & $7.7$ & $35.7$ \\ \hline 
				\multicolumn{4}{|c|}{\textit{Subset Percentage 30\%}} \\ \hline
				PMLP-Random & $23.9$ & $12.4$ & $47.7$ \\ \hline 
				PMLP-Top-Loss & $17.5$ & $7.8$ & $33.9$ \\ \hline 
				PMLP-C-Top-loss & $19.50$ & $9.7$ & $38.0$ \\ \hline  
			\end{tabular}
		}
	\end{center}
\end{table}

To evaluate the effectiveness of the proposed subset sampling and online hyper-parameter selection method, we perform experiments on publicly available datasets designed for object recognition (Caltech256 \cite{griffin2007caltech}), indoor scene recognition (MIT \cite{quattoni2009recognizing}) and face recognition (CelebA \cite{liu2015deep}) problems. For CelebA dataset, we used a subset of $60000$ images corresponding to $500$ identities. In each dataset, $80\%$, $10\%$ and $10\%$ of the data were used for training, validation and testing, respectively. The inputs to all PNNL algorithms are deep features (global average pooling of the last convolution layer) from a pre-trained network on ImageNet dataset \cite{simonyan2014very}. We demonstrate subset sampling with subset percentage of $10\%$, $20\%$, $30\%$, and online hyper-parameter selection on PMLP. As previously mentioned, the adopted PMLP follows the progression rule of HeMLGOP in \cite{tran2019heterogeneous}. Here we should note that subset selection was used to only speed-up the network progression process (topology construction); the final topologies were fine-tuned with the full set of training data. We also evaluated other PNNL algorithms, namely StackedELM \cite{zhou2014stacked} and PLN \cite{chatterjee2017progressive} which run on the full training set at each step. For detailed information about our experiment protocols and hyper-parameter setting, we refer the readers to our publicly available implementation of this work\footnote{https://github.com/viebboy/SIPL}.

 \begin{table}[t!]
	\begin{center}
		\caption{Average time taken to optimize one block of neurons (in seconds)}\label{t3}
		\resizebox{0.7\linewidth}{!}{
			\begin{tabular}{|c|c|c|c|}
				\multicolumn{4}{c}{} \\ \hline
				\textbf{Models}		& \textbf{Caltech256} 	& \textbf{MIT} & \textbf{CelebA} \\ \hline \hline
				\multicolumn{4}{|c|}{\textit{Subset Percentage 10\%}} \\ \hline
				PMLP-Random & $6.7$ & $3.4$ & $20.1$ \\ \hline 
				PMLP-Top-Loss & $5.6$ & $2.7$ & $10.8$ \\ \hline 
				PMLP-C-Top-Loss & $8.11$ & $3.0$ & $17.6$ \\ \hline \hline
				\multicolumn{4}{|c|}{\textit{Subset Percentage 20\%}} \\ \hline
				PMLP-Random & $10.3$ & $5.5$ & $27.8$  \\ \hline
				PMLP-Top-Loss & $6.6$ & $4.14$ & $18.9$ \\ \hline
				PMLP-C-Top-loss & $12.6$ & $4.3$ & $35.7$ \\ \hline \hline
				\multicolumn{4}{|c|}{\textit{Subset Percentage 30\%}} \\ \hline
				PMLP-Random & $12.4$ & $6.3$ & $34.5$ \\ \hline
				PMLP-Top-Loss & $12.6$ & $4.8$ & $38.6$ \\ \hline
				PMLP-C-Top-loss & $15.1$ & $5.5$ & $44.7$ \\ \hline \hline
				\multicolumn{4}{|c|}{\textit{Full Set}} \\ \hline
				PMLP & $72.2$ & $30.5$ & $170.3$ \\ \hline
				StackedELM \cite{zhou2014stacked} & $2.9$ & $1.5$ & $8.7$ \\ \hline
				PLN \cite{chatterjee2017progressive} & $113.7$ & $17.2$ & $528.4$ \\ \hline	 
			\end{tabular}
		}
	\end{center}
\end{table}
Table \ref{t1} shows the recognition accuracy on the test set of all models on the three datasets. For compact presentation, we refer to the proposed PMLP variants based on \textit{Random Sampling}, \textit{Top-M Sampling based on miss-classification} and \textit{Top-M Sampling based on diverse miss-classification} by PMLP-Random, PMLP-Top-Loss and PMLP-C-Top-Loss, respectively. Different from the empirical results obtained in other learning contexts, the best performing subset selection strategy is random sampling at the lowest percentage level ($10\%$). In fact, PMLP-Random at $10\%$ performs better than all other algorithms, including the original PMLP. This can be attributed to the effects of both random subset sampling and online hyper-parameter selection. Random sampling with a small percentage leads to the general effect that different blocks of neurons are optimized with respect to diverse subsets of data. The final network after optimization can be loosely seen as an ensemble of smaller networks. On the other hand, when a subset of data persists being miss-classified throughout the network's topology progression process, the corresponding sampling strategies will bias the algorithm to select only these samples and reduce the diversity of information presented to the network.

Table \ref{t2} shows the total number of unique samples selected by each algorithm following different sampling strategies. This table reflects the degree of diversity in the inputs observed by different networks trained with different sampling strategies. It is clear that the numbers are much higher for random sampling. While the original PMLP presents greater amount of information to the network during progression, every block of neurons in PMLP observes the same set of data, which might lead to over-fitting. 

Table \ref{t3} shows the average time taken to optimize one block of neurons in each algorithm while Table \ref{t4} shows the total time taken to perform experiments for a particular setting. Every experiment run was performed on the same node configuration (4 CPU cores, 16 GB of RAM). It is clear that using subset selection, the average time taken at each step of PMLP is greatly reduced (Table \ref{t3}). Combining subset selection and online hyper-parameter selection, the total experiment time is significantly lower (Table \ref{t4}). 

\begin{table}[t!]
	\begin{center}
		\caption{Total time taken to perform all experiments (in hours)}\label{t4}
		\resizebox{0.7\linewidth}{!}{
			\begin{tabular}{|c|c|c|c|}
				\multicolumn{4}{c}{} \\ \hline
				\textbf{Models}		& \textbf{Caltech256} 	& \textbf{MIT} & \textbf{CelebA} \\ \hline \hline
				\multicolumn{4}{|c|}{\textit{Subset Percentage 10\%}} \\ \hline
				PMLP-Random & $5.2$ & $3.1$ & $14.5$ \\ \hline
				PMLP-Top-loss & $6.5$ & $3.6$ & $18.9$  \\ \hline
				PMLP-C-Top-Loss & $7.9$ & $4.0$ & $23.8$ \\ \hline \hline
				\multicolumn{4}{|c|}{\textit{Subset Percentage 20\%}} \\ \hline
				PMLP-Random & $5.2$ & $2.6$ & $14.4$ \\ \hline
				PMLP-Top-loss & $6.5$ & $3.6$ & $18.3$ \\ \hline
				PMLP-C-Top-loss & $8.6$ & $3.8$ & $24.7$ \\ \hline \hline
				\multicolumn{4}{|c|}{\textit{Subset Percentage 30\%}} \\ \hline
				PMLP-Random & $5.4$ & $2.8$ & $14.8$ \\ \hline
				PMLP-Top-loss & $7.0$ & $3.6$ & $19.5$ \\ \hline
				PMLP-C-Top-loss & $7.8$ & $4.1$ & $24.8$ \\ \hline \hline
				\multicolumn{4}{|c|}{\textit{Full Set}} \\ \hline
				PMLP & $18.4$ & $6.0$ & $56.3$ \\ \hline
				StackedELM \cite{zhou2014stacked} & $0.19$ & $0.17$ & $0.3$ \\ \hline
				PLN \cite{chatterjee2017progressive} & $862.1$ & $100.0$ & $5185$ \\ \hline	 
			\end{tabular}
		}
	\end{center}
\end{table}

\section{Conclusion}
In this work, we proposed subset sampling and online hyper-parameter selection to speed up and enhance PNNL algorithms. Empirical results demonstrated with PMLP show that proposed approach can not only accelerate the optimization procedure in PMLP but also improve the generalization performance of the resulting networks. 

\section{Acknowledgement}
This project has received funding from the European Union's Horizon 2020 research and innovation programme under grant agreement No 871449 (OpenDR). This publication reflects the authors’ views only. The European Commission is not responsible for any use that may be made of the information it contains.

\bibliography{reference}
\bibliographystyle{ieeetr}

\end{document}